\documentclass[journal]{IEEEtran}
\ifCLASSINFOpdf
\else
\fi
\usepackage[dvipsnames]{xcolor}
\usepackage{times}
\usepackage{epsfig}
\usepackage{graphicx}
\usepackage{amsmath}
\usepackage{amssymb,adjustbox,ctable,threeparttable,booktabs,mathrsfs,verbatim,algorithm,algpseudocode}

\usepackage[pagebackref=true,breaklinks=true,letterpaper=true,colorlinks,bookmarks=false]{hyperref}

\newcommand{\etal}{\textit{et al}. }

\makeatletter
\newcommand\footnoteref[1]{\protected@xdef\@thefnmark{\ref{#1}}\@footnotemark}
\makeatother

\begin{document}
%
\title{End-to-End Pore Extraction and Matching in Latent Fingerprints: Going Beyond Minutiae}

%
%
%

\author{Dinh-Luan~Nguyen
        and~Anil K.~Jain
\thanks{Dinh-Luan Nguyen and Anil K. Jain are with the Department
of Computer Science and Engineering, Michigan State University, East Lansing,
MI, 48824 USA. \protect\\
E-mail: nguye590@msu.edu and jain@cse.msu.edu.}
}

%
%


\markboth{}%
{}

%



\maketitle

\begin{abstract}
Latent fingerprint recognition is not a new topic but it has attracted a lot of attention from researchers in both academia and industry over the past 50 years. With the rapid development of pattern recognition techniques, automated fingerprint identification systems (AFIS) have become more and more ubiquitous. However, most AFIS are utilized for live-scan or rolled/slap prints while only a few systems can work on latent fingerprints with reasonable accuracy. The question of whether taking higher resolution scans of latent fingerprints and their rolled/slap mate prints could help improve the identification accuracy still remains an open question in the forensic community. Because pores are one of the most reliable features besides minutiae to identify latent fingerprints, we propose an end-to-end automatic pore extraction and matching system to analyze the utility of pores in latent fingerprint identification. Hence, this paper answers two questions in the latent fingerprint domain: (i) does the incorporation of pores as level-3 features improve the system performance significantly? and (ii) does the 1,000 ppi image resolution improve the recognition results? We believe that our proposed end-to-end pore extraction and matching system will be a concrete baseline for future latent AFIS development.

\end{abstract}


%
\IEEEpeerreviewmaketitle

\section{Introduction}
%
%
%
%
\IEEEPARstart{L}{atent} fingerprints are unintentional fingerprints that are left on surfaces. They are challenging to identify because of many factors such as poor quality, background noise, distortion, small non-overlapping areas, etc. Hence, processing latent images requires a lot of effort and time to get acceptable recognition results compared to scanned or rolled fingerprints. Unlike rolled fingerprints which have a number of distinctive characteristics such as ridge flow, large number of minutiae, and little background noise, latent fingerprints only contain a small portion of the fingerprint ridge structure \cite{nguyen2018robust, nguyen2019automatic}. Thus, there is an urgent need to use other features that are present even in small partial fingerprints. 

Fingerprints are known to have distinctive characteristics, even in identical twins, and are unchanging throughout a person's lifetime \cite{jain2007pores}. Such discriminative power emerges from three levels of fingerprint features: (i) Level 1 (ridge flow); (ii) Level 2 (minutiae points, such as ridge bifurcations and endings); and (iii) Level 3 (scars, pores, incipient ridges, etc.). Automatic fingerprint identification systems (AFIS) have generally used minutiae as primary features to discriminate fingerprints from different subjects. 
While AFIS accuracy for matching full fingerprints to full fingerprints is sufficiently high \cite{indovina2011elft}, only a few commercial AFIS can work on latent images and even then with low accuracy. The reason is that most of AFIS primarily rely on minutiae for matching while many of latent images just contain a small part of a fingerprint, and thus few minutiae. As a result, the AFIS fails in those cases. In developing a robust latent AFIS, one should take care to handle the cases where the query latent contains only a small portion of the whole fingerprint ridge flow. Use of level 3 features is a natural solution for such scenarios.

Unlike scanned or rolled fingerprints, latent fingerprints are very sensitive to both inherent (e.g. scars, injuries) and background (e.g. friction ridges on a printed document) noise. Since sweat pores are on fingerprint ridges are quite visible to human eyes compared to the other level-3 features, pores have been used as complementary features to minutiae \cite{zhao2010latent, jain2007pores, zhao2009direct}. This argument is the basis for our usage of pores as main level-3 features for high resolution latent fingerprint matching. If successful, the use of pores for matching could be used in other applications domains involving partial fingerprints (due to small sensor size), such as smart bank cards and mobile devices. Pores could also be used to aid infant fingerprint recognition systems where minutiae are not always available, even at resolutions over 1500 ppi.

\begin{figure*}[!tbp]
\centering
\includegraphics[width=\textwidth]{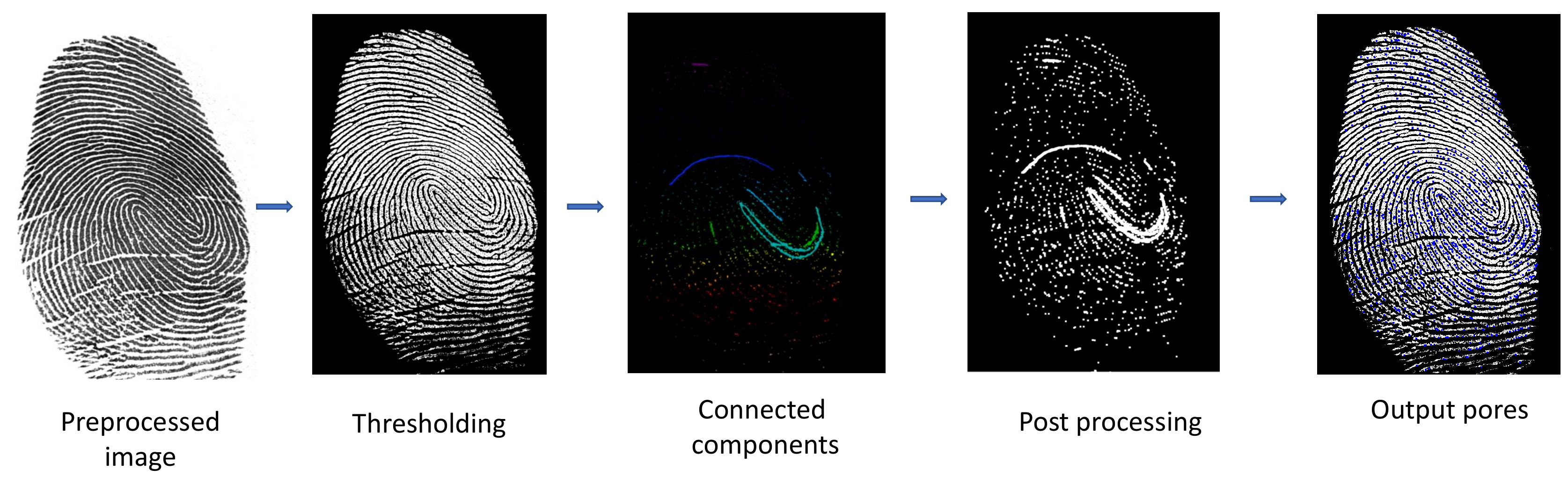}
\caption{Pore extraction pipeline for any pre-processed input image. There are four steps: (i) apply an adaptive thresholding technique with morphology operations to remove noise; (ii) extract connected components having sizes similar to pores; (iii) eliminate large and non-round shaped components; (iv) output set of coordinates representing relative pore locations.}
\label{fig:PoreExtraction}
\end{figure*}

\section{Related Work}
An end-to-end AFIS contains two stages: feature extraction and matching. The feature extraction stage converts input fingerprint images into corresponding templates while the matching stage uses those templates to compute a similarity score between the input fingerprint templates, resulting in a ranked list of candidate matches. Although many studies have tried to address level 3 feature extraction and matching on fingerprints, these only used high resolution images with no background noise \cite{zhao2009direct} or scanned fingerprints. 

\subsection{Pore extraction.} 
With the advancement of capture technology, obtaining high resolution images is not an obstacle in forensic community. One of the pioneers in the extraction of pores as fingerprint features is Ray \etal \cite{ray2005novel}, who used a modified minimum squared error approach on the NIST 4 \cite{NSITDB4} dataset. Jain \etal \cite{jain2007pores} developed their techniques based on high quality scanned partial fingerprint images. Zhao \etal \cite{zhao2010adaptive} proposed adaptive modeling for sweat pore extraction. They divided the image into small blocks and constructed local pore models for each region.

With the development of neural networks and high-powered GPUs, some studies have used a deep network approach to extract pores. Jang \etal \cite{jang2017deeppore} proposed a pore extraction technique based on convolutional neural networks. Again, they used high resolution partial fingerprints, not latents, to evaluate their approach. Genovese \etal \cite{genovese2016towards} utilized a neural network to extract pores from live-scanned fingerprints. Labati \etal \cite{labati2018novel} used a similar idea for pore extraction in heterogeneous fingerprint images. However, the lab-collected ``pseudo latent'' dataset mentioned in the paper is very different from those sourced from forensic agencies.

The excellent pore extraction results on scanned high resolution images with clear backgrounds do not reveal anything about their capability on operational latent fingerprint datasets. Besides, for rolled or live-scanned high-resolution fingerprint images, traditional approaches can achieve reasonable results.

\subsection{Pore matching} 
There have been several attempts made to use pores as a feature for matching. Jain \etal \cite{jain2007pores} designed a hand-crafted approach to utilize features from all 3 fingerprint feature levels. They used pre-defined thresholds for their matching system. Again, they only reported experimental results on their high-resolution lab-collected dataset. Chen and Jain \cite{chen2007dots} extracted an extended feature set as input for the matching step. They used a commercial off-the-shelf (COTS) matcher to evaluate their feature extraction. The dataset on which they evaluated is the partial fingerprint dataset from NIST SD30 \footnote{\label{note1}This dataset is no longer publicly available.}. Zhao \etal \cite{zhao2009direct} proposed a pore matcher based on a combination of the ICP \cite{rusinkiewicz2001efficient} and RANSAC \cite{fischler1981random} algorithms. Again, they only evaluated their algorithm on their high-resolution live-scanned partial fingerprint images. 

A few publications applied their algorithms on latent fingerprint images. Jain and Feng \cite{jain2011latent} combined pores with other manually marked features for matching. However, their approach is not fully automated, requiring human intervention for feature extraction. The work in \cite{zhao2010latent} evaluated directly on latent fingerprint images using level-3 features. They also used other level 3 features such as dots, incipient ridges, and ridge edge protrusions (DIP) for the matching. However, their approach was also not fully automatic and relied on an existing COTS minutiae matcher to see the advantage of using extended feature sets. Although their approach can work reasonably well on simulated partial fingerprints, it failed when used independently without a COTS system. Other existing pore matching works (\cite{dahia2018cnn, liu2017feature}) were not applied or evaluated on latent images, or simply do not go beyond the pore extraction stage \cite{zhao2009direct, jang2017deeppore, labati2018novel}. One of the reasons for the scarcity of latent fingerprint research using pores is the lack of publicly available operational latent fingerprint databases. 

\begin{figure*}[!tbp]
\centering
\includegraphics[width=\textwidth]{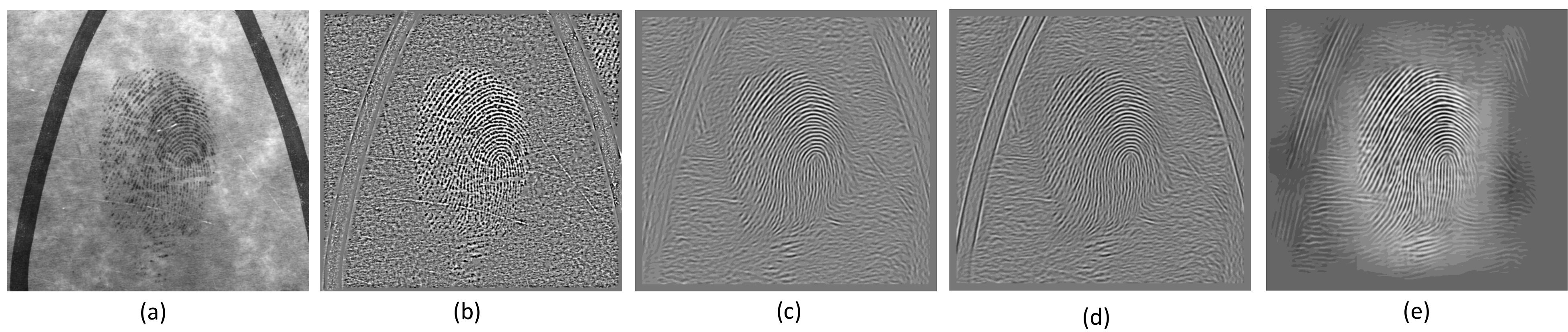}
\caption{Four examples of latent fingerprint enhancement: (a) Input latent; (b) STFT; (c) Cartoon-texture based; (d) Contrast enhancement; (e) Auto-encoders. Those enhancement images convert the latent and rolled prints into a similar appearance, thus projecting them into the same input space and allowing for easier extraction of connected components.}
\label{fig:Enhancement}
\end{figure*}

\section{Pre-processing and pore extraction}
There are three reasons, in our opinion, for not using a neural network approach to extract pores in forensic latent fingerprint images: (i) latent fingerprints contain unexpected background noise and distortion; (ii) pores in latent fingerprints are very sensitive to image normalization techniques; and (iii) pores are very small compared to the overall image size, which are not suitable for object detection-like neural network approaches.

\begin{figure*}[!tbp]
\centering
\includegraphics[width=\textwidth]{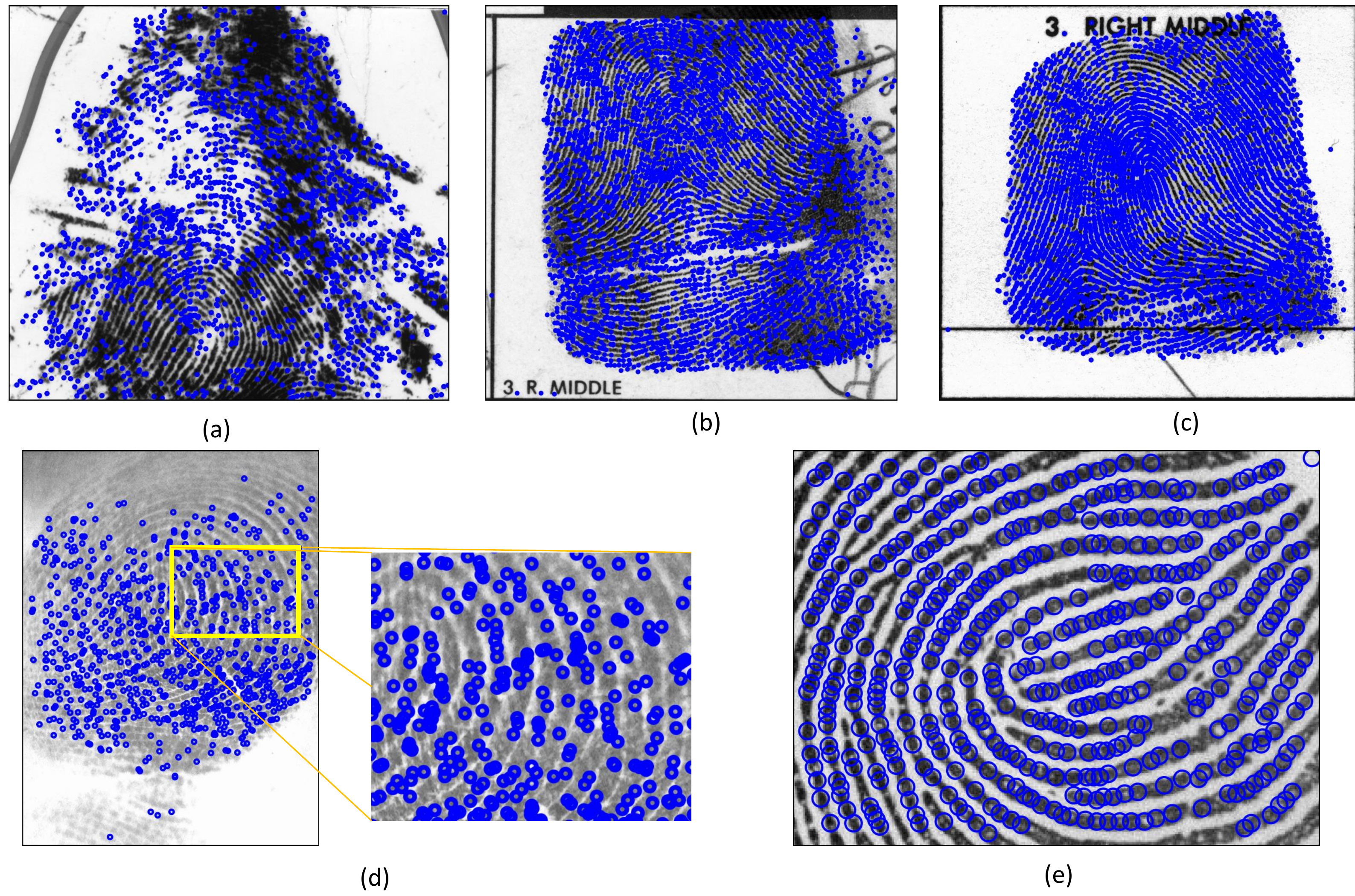}
\caption{Pore extraction results of different fingerprint types and resolutions: (a) Latent fingerprint from NIST SD27 (1,000 ppi); (b) Rolled prints in NIST SD27 (500 ppi); (c) Rolled prints in NIST SD30 (1,000 ppi); (d) Infant fingerprint (1,900 ppi); (e) Partial fingerprint from PolyU-HRF \cite{polyuHRF} (1,200 ppi)}
\label{fig:PoreExample}
\end{figure*}

\subsection{Pore Extraction}
Unlike other pore extraction methods using mathematical modeling \cite{zhao2009direct, zhao2010latent} or pixel values between ridge and valley \cite{jain2007pores} that are only suitable for images with clear background, our proposed approach can work on arbitrary input fingerprint images without remodeling for new fingerprint image type or resolution. Based on the observation that sweat pores are on the fingerprint ridges, we propose two stages for pore extraction: (i) ridge enhancement and (ii) pore extraction using connected components. Figure \ref{fig:PoreExtraction} shows the pipeline of our proposed pore extraction module. Note that we visualize pore extraction steps on high-resolution scanned images to allow readers to understand the approach more easily.

\subsubsection{Fingerprint ridge enhancement}
\label{sec:enhancement}

Unlike scanned fingerprints which have a clear background, latent fingerprints need to be enhanced to obtain reliable friction ridges for further processing. We use several different enhancement techniques to get enhanced images for a given latent fingerprint input. Specifically, we apply the short-time Fourier transform (STFT) \cite{chikkerur2007fingerprint} and cartoon-texture decomposition technique \cite{buades2011cartoon+} to enhance the the original image and get sharper friction ridges. Since pores have small sizes and variable pixel values, global enhancement techniques such as contrast enhancement \cite{stark2000adaptive} or auto-encoders \cite{cao2018end} are not suitable here. This is in contrast to systems that use minutiae or orientation field as features, where these enhancement techniques are appropriate. Figure \ref{fig:Enhancement} shows examples of latent fingerprint enhancement. Although the STFT and cartoon-texture decomposition enhancement approaches do not completely eliminate background noise, they are resistant to artifact creation, and can be used to convert the latent and rolled prints into a similar appearance, thus projecting them into the same input space and allowing for easier extraction of connected components. Hence, the system does not need to have different configurations for different types of input images.

\begin{figure*}[!tbp]
\centering
\includegraphics[width=\textwidth]{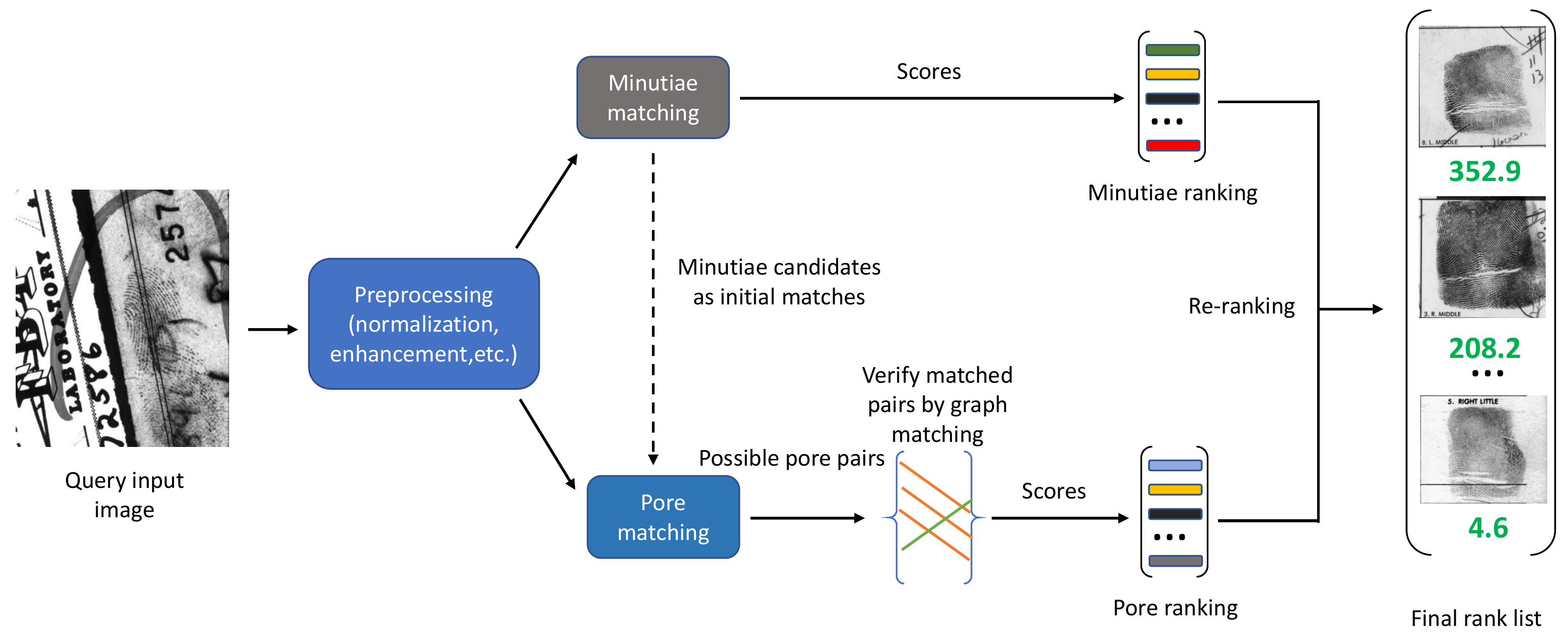}
\caption{Graph matcher pipeline. There are two main components: minutiae-based vs. pore-based matching, and re-ranking. Minutiae are utilized to get initial matches for the pore-based matcher. Correspondences are verified using bipartite graph matching. The final rank list is the result of re-ranking the minutiae candidate list based on pore matching scores.}
\label{fig:Matching}
\end{figure*}

\subsubsection{Neighbors connected component technique for pore extraction}

Since all fingerprint (latent, rolled, slap, or scanned) images are projected to a single input space based on the approach in section \ref{sec:enhancement}, we apply the adaptive threshold technique \cite{bradley2007adaptive} to get a binary version of the image for the extraction of connected components. Because pores are local features, using the adaptive threshold technique can divide the input image into sub-regions and binarize that region based on its information. Thus, this technique can preserve location of pores.

Based on the observation that pores are small, have rounded shape and are located on the friction ridges, we regard the image as a densely connected graph where an edge exists between two vertices if and only if they have the same pixel color. The problem turns out to be the graph coloring problem \cite{jensen2011graph}, which is easily solved using graph traversal and some special data structures. We refer readers to \cite{pardalos1998graph} for more information about the solution to this problem. Because pores are small in size and vary based on input resolution, we create a user-input parameter $P$ for limiting the maximum pore area. By default, $P=100$ pixels is used for $1,000$ ppi image. Thus, components having size larger than $P$ are eliminated. The precise location of each pore is taken to be the mean of all pixels inside that pore region.

Figure \ref{fig:PoreExample} shows pore extraction results on different fingerprint types and resolutions to demonstrate the generalizability of our proposed pore extraction approach.


\begin{figure*}[!tbp]
\centering
\includegraphics[width=\textwidth]{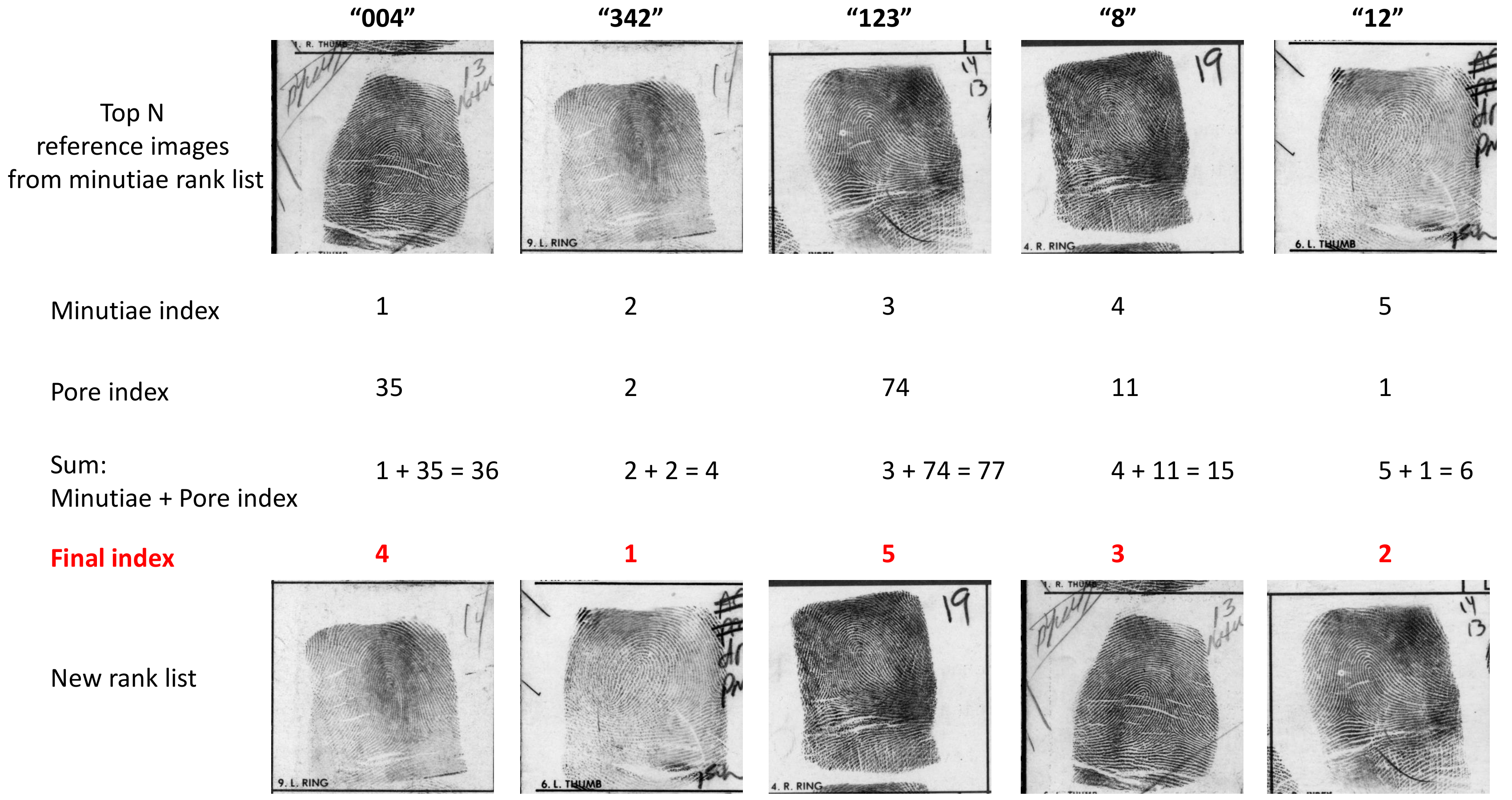}
\caption{Illustration of pore re-ranking $N=5$. New index for each candidate image in minutiae rank list is calculated as the sum of minutiae and pore index.}
\label{fig:Rerank}
\end{figure*}

\section{Pore Matching With Graphs}
Figure \ref{fig:Matching} visualizes our matching framework. Our approach comprises of two modules: minutiae vs. pore matching and re-ranking. 
Since pores are only used for exclusion in cases where the minutiae decision is not clear, we use pore matching as a complementary step for minutiae matching. Hence, given an input latent image, we normalize its resolution to $500$ ppi, apply our enhancement technique given in section \ref{sec:enhancement}, and input it to an end-to-end minutiae-based matcher \cite{cao2018end}. There are two cases in which the minutiae matcher does not provide strong discriminative power: (i) number of minutiae is small; and (ii) similarity scores between matched minutiae are not high enough (i.e. weak confidence) to deliver a decision.

\begin{table*}[!tbp]
	\centering
	\caption{Summary of NIST and a public fingerprint databases.}    
	\label{tab:database}
	\begin{normalsize}
	\begin{adjustbox}{max width=\textwidth}
		\begin{tabular}{c|c|c|c}
			\hline
			{Database}& {Number of images}& {Source}& {Note}\\
			\hline
			\hline
			{NIST SD27}	&255&Forensic agency&1,000 ppi Latent database\\
			{NIST SD30}	&360&Forensic agency&Rolled prints at both 500 and 1,000 ppi\\
			{PolyU-HRF}	&1,480&Laboratory&Partial rolled prints at 1,200 ppi\\
			\hline
		\end{tabular}
		\end{adjustbox}
	\end{normalsize}
\end{table*}

\subsection{Pore matching using bipartite graphs}
Before discussing the details of pore matching, we summarize the common steps for minutiae matching: (i) detect minutiae; (ii) construct minutiae descriptors using surrounding patches around minutiae by hand-crafted approaches or automatically extracted from convolutional networks; (iii) find minutiae correspondences by comparing distances (e.g. $L_1$ or $L_2$ norm, cosine criteria, etc.) between minutiae descriptors; (iv) eliminate wrong (low confidence) matches using relative position between minutiae; (v) return the final score as the sum of all minutiae similarity scores \cite{cao2018end}. Our proposed pore matching approach follows this procedure to find reliably matched pores. Note that we do not use a neural network approach to get pore descriptors (as in the construction of minutiae descriptors) because pores are dense and small, which contrasts with the sparse and relatively large number of attributes of minutiae. Since the surrounding area for each pore is small (i.e. $10 \times 10$ pixels in $1,000$ ppi image), they are not suitable for deep networks.
 
Unlike scanned/rolled fingerprints, latent fingerprints usually have a small number of minutiae, and thus cannot be reliably matched using minutiae alone. The average number of minutiae in NIST SD27 is $13$ and the minimum and maximum number of minutiae are: $1$ and $63$ On the other hand, the average number of minutiae in rolled mates of NIST SD27 latents is $58$.
However, that does not mean we completely disregard the minutiae information. On the contrary, we use matched minutiae to guide initial pore correspondence. To increase system's ability to handle extreme cases (fewer than $3$ matched minutiae), we consider two cases: fewer than $3$ minutiae matches, and at least $3$ minutiae matches. Specifically, given at least $3$ matched minutiae pairs, we construct a transformation matrix from a latent fingerprint to a rolled print. Since different impressions are not collected at the crime scenes in the same way, we introduce $\Delta$ tolerance to deal with distortion. That means one latent pore can have a few matched pores in a given rolled print which satisfy the tolerance constraint. In the case of fewer than $3$ matched minutiae, we construct a Hamiltonian path \cite{gurevich1987expected} on both the latent print and rolled print pore graphs. Then, we apply statistical shape matching \cite{castellani2008sparse} to get matched pore pairs.

At this point, each pore in the latent print has a few candidate pores in the rolled print. We remove spurious matches by converting this problem into bipartite matching: finding the maximum set of matches in such a way that no two matches have a common endpoint. Because the algorithm finds the maximize number of matches, only very similar image pairs can have high similarity scores (as well as number of matches), where the pore matching score between two images is the sum of scores of all refined pore matches. 

\subsection{Re-ranking with fusion}
Since pores, in forensic practice, are not used directly for identification (instead for exclusion), we use pore scores as criteria to re-rank the minutiae matching-based rank list. Figure \ref{fig:Rerank} presents our re-ranking process. First, we get pore scores for top $N$ candidates in the minutiae rank list. Second, we retrieve the pore index for each pore score. Third, we compute the weighted sum of minutiae and pore indices. Lastly, the final index is the ranking of the sum calculated in the previous step.

\section{Experiments}
\begin{figure*}[!tbp]
\centering
\includegraphics[width=\textwidth]{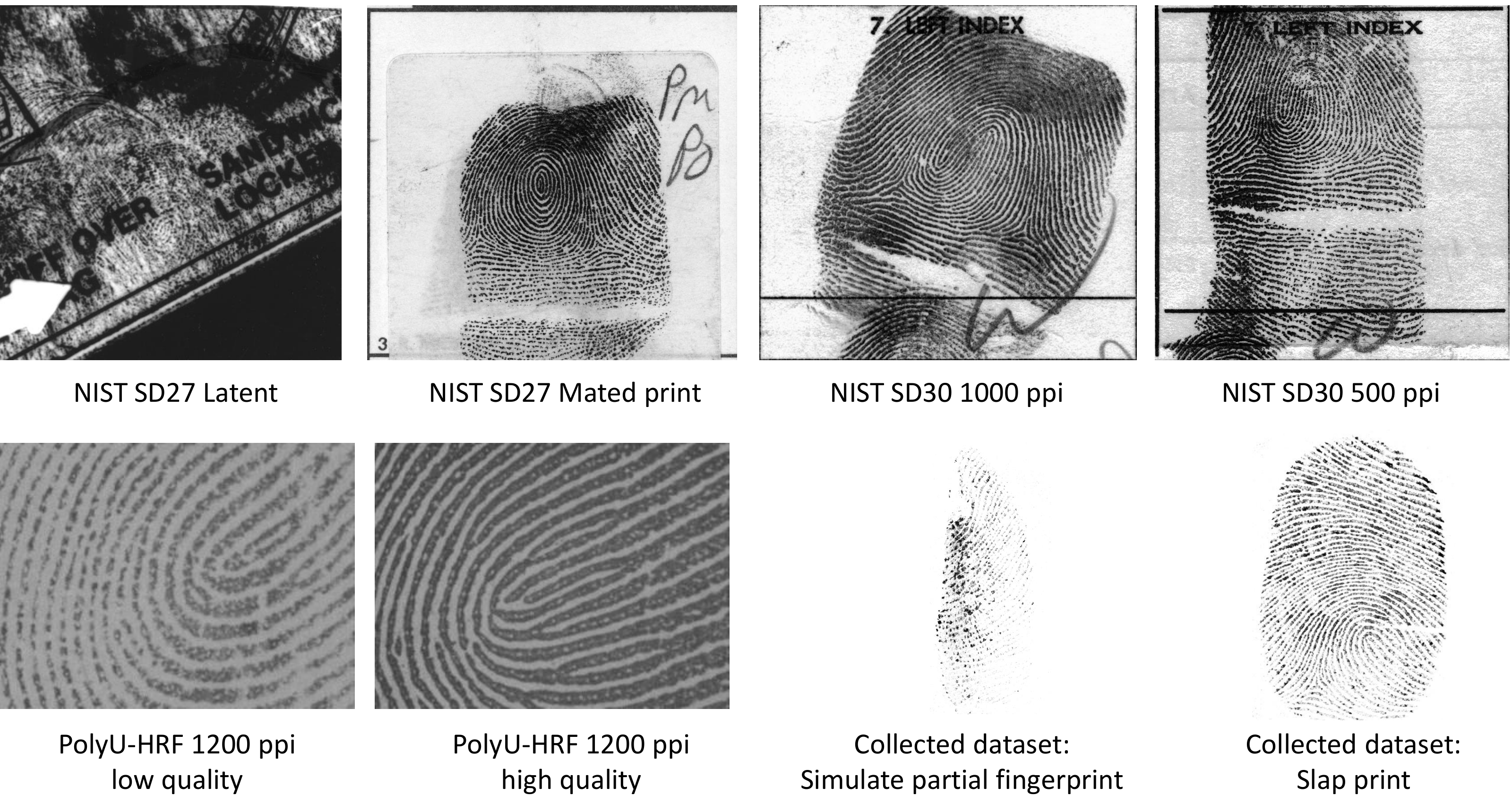}
\caption{Examples of fingerprint images from NIST, PolyU-HRF, and a lab collected database.}
\label{fig:ExampleDatabase}
\end{figure*}

In this paper, three databases, NIST SD27 (1,000 ppi)\footnoteref{note1}, NIST SD30\footnoteref{note1}, and PolyU-HRF \cite{polyuHRF}, are used to evaluate the proposed end-to-end pore extraction and matching framework. Table \ref{tab:database} shows a summary of the three fingerprint databases. We also collected a database of 500 different fingers from 50 subjects at both 500 ppi and 1,000 ppi for our algorithm development and parameter tuning. Some example images from the NIST, PolyU-HRF, and our privately collected databases are shown in figure \ref{fig:ExampleDatabase}.

\subsection{Pore extraction evaluation}
As shown in figure \ref{fig:PoreExample}, our proposed pore extraction approach can work well on a variety of input image types and resolutions. However, with the scarcity of high-resolution fingerprint datasets with ground truth information on pores, we quantitatively evaluate the effectiveness of our approach on $30$ partial fingerprint images \cite{polyuHRF} for reference only. Note that our algorithm's better pore extraction ability on non-latent high-resolution fingerprint images does not predict its capability on latent fingerprint images. 

\subsubsection{Evaluation protocol}
We use the precision, recall, and F1 score to evaluate the goodness of extracted pores compared to manual ground truth\footnote{Provided by the authors of \cite{zhao2010latent}}. Let $TP$ be true positive, $FP$ be false positive, $FN$ be false negative. Precision, recall, and F1 score are then defined as:

\begin{align}
\begin{split}
Precision = \frac{TP}{TP+FP};  
\\
Recall = \frac{TP}{TP+FN};
\\
F1 = 2\times \frac{Precision \times Recall}{Precision + Recall}
\end{split}
\end{align}

Table \ref{tab:quanExtraction} shows the precision, recall, and F1 score of pore extraction on partial fingerprint images. The default confident threshold is $50$. All of the precision, recall, and F1 score are in the range of $[0,1]$ where value of $1$ indicates the ``perfect'' compared to the ground truth. 

\begin{table}[!tp]
	\centering
	\caption{Quantitative pore extraction evaluation of our proposed approach on partial fingerprint database.}    
	\label{tab:quanExtraction}
	\begin{small}
	\begin{adjustbox}{max width=\textwidth}
		\begin{tabular}{c|c|c|c}
			\hline
			{Threshold of confidence}& {Precision}& {Recall}& {F1 score}\\
			{in the binarization step}& {}& {}& {}\\
			\hline
			\hline
			{30}	&$47.8\%$&$28.2\%$&$0.355$\\
			{50}	&$73.9\%$&$43.7\%$&$0.549$\\
			{70}	&$65.3\%$&$49.5\%$&$0.563$\\
			\hline
		\end{tabular}
		\end{adjustbox}
	\end{small}
\end{table}

\begin{figure}[!tbp]
\centering
\includegraphics[width=\columnwidth]{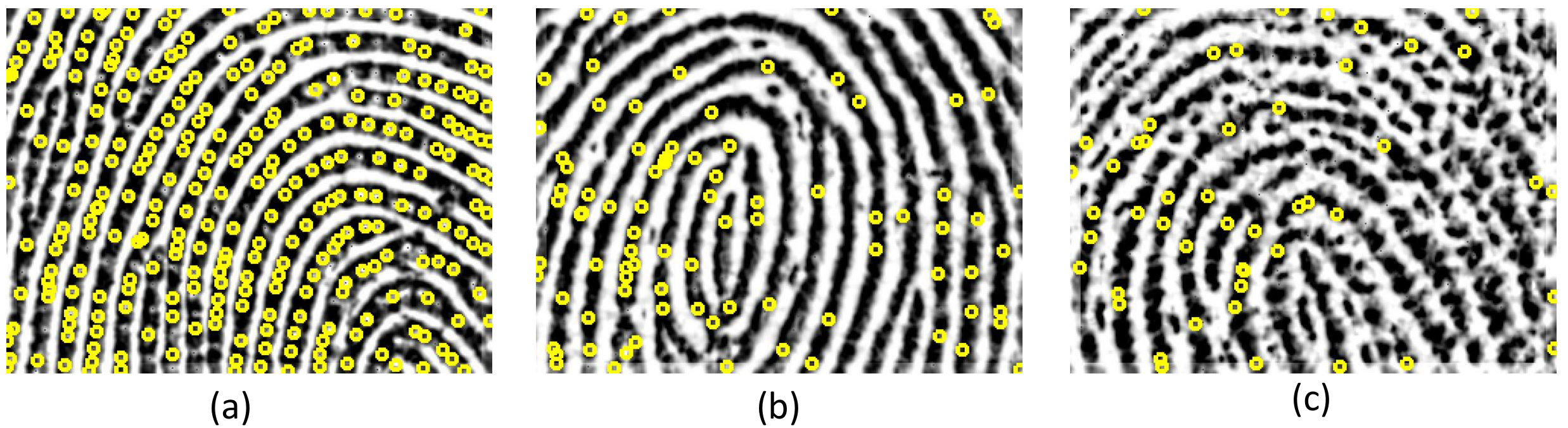}
\caption{Pore extraction on partial fingerprints with different qualities: (a) high; (b) normal; and (c) low.}
\label{fig:PartialFingerprint}
\end{figure}

\begin{figure}[!tbp]
\centering
\includegraphics[width=\columnwidth]{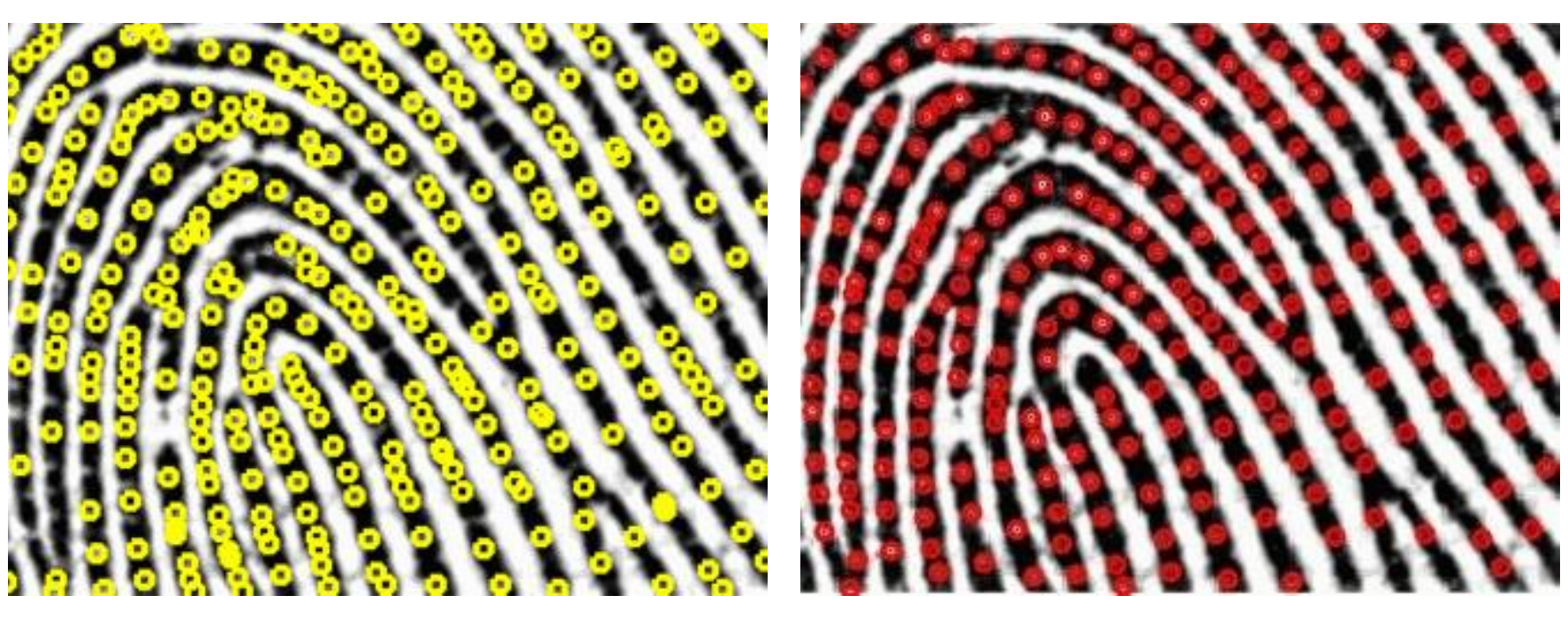}
\caption{Visual comparison of our pore extraction results (left) and manual ground truth (right).}
\label{fig:CompareFingerprint}
\end{figure}

Figure \ref{fig:PartialFingerprint} and figure \ref{fig:CompareFingerprint} show pore extraction on different image qualities (from high to low), and visual comparison between our proposed pore extraction and manual ground truth, respectively. Our proposed approach performs fairly well, especially on low quality fingerprint images.

\subsubsection{Running time}
The pore extraction time depends on the number of pores in the input image. In the NIST SD27 1,000ppi dataset, the latent images, on average, take $0.03 \pm 0.015$ seconds for extraction, while the rolled prints (500 ppi upsampled to 1,000 ppi) take $0.57 \pm 0.017$ seconds on an Intel(R) Core(TM) i7-7700 CPU @3.6GHz, 32GB RAM desktop machine. Approximately, the computational complexity of pore extraction is an exponential function of the number of pores.

\subsection{Pore matching evaluation}
\begin{figure}[!tbp]
\centering
\includegraphics[width=\columnwidth]{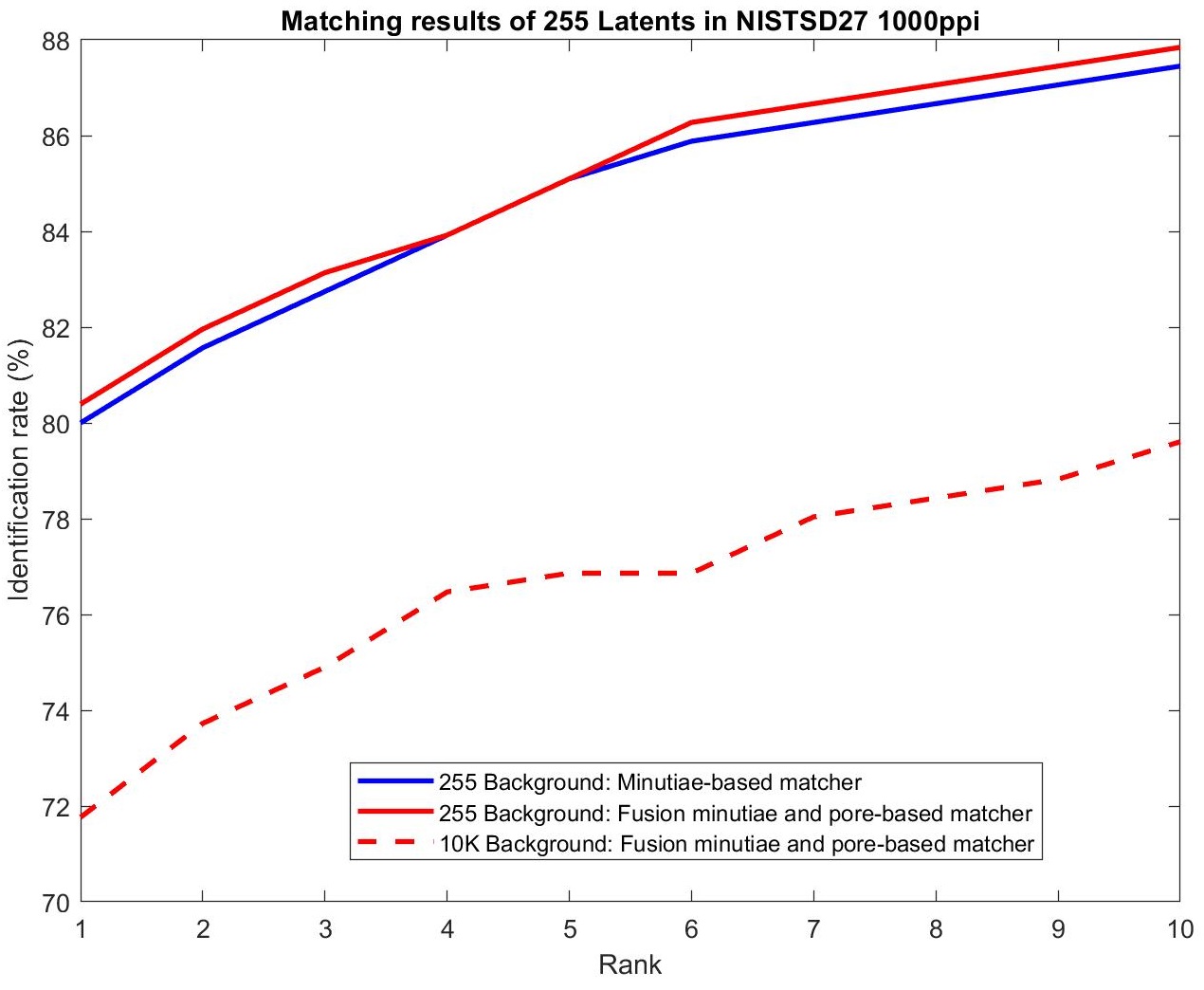}
\caption{CMC curves of our pore based matcher on $255$ latents in NIST SD27 1,000 ppi with background size up to $10K$. All 500 ppi background images are upsampled to 1,000 ppi in our experiment. For the $10K$ background, the curves of the minutiae matcher and the proposed approach involving pore matching completely overlap.}
\label{fig:CMC10k}
\end{figure}

We evaluate our proposed pore-based matcher on the NIST SD27 1,000 ppi dataset. We report the performance based on closed-set identification, where the true mated prints of the query are assumed to be in the gallery. Figure \ref{fig:CMC10k} shows the cumulative match characteristic (CMC) curves of our proposed pore-based matcher and minutiae-based matcher on $255$ latents in NIST SD27 1,000 ppi with background size up to $10,000$.

By utilizing pore information, we improve $3$ more cases in rank-1 and $9$ cases at rank-10 compared to using minutiae matcher only. Thus, the overall rank-1 accuracy is $80.4\%$ with a background size of 255. This implies that pores provide some information that is complementary to the minutiae. However, when the background size increases, the pore based matcher no longer improves the minutiae based matcher and even decrease the rank retrieval for some of the latents. The reason for showing only a single curve for $10,000$ background is that the pores do not help minutiae, hence the minutiae matcher curve and fusion curve are the same. That means there are no cases which satisfy the pore matching criteria (e.g. pore scores are not significantly distinct or the minutiae scores between latent and imposter prints are already sufficiently large to make a decision). Further explanation is provided in section \ref{sec:analysis}.

\begin{figure}[!tbp]
\centering
\includegraphics[width=\columnwidth]{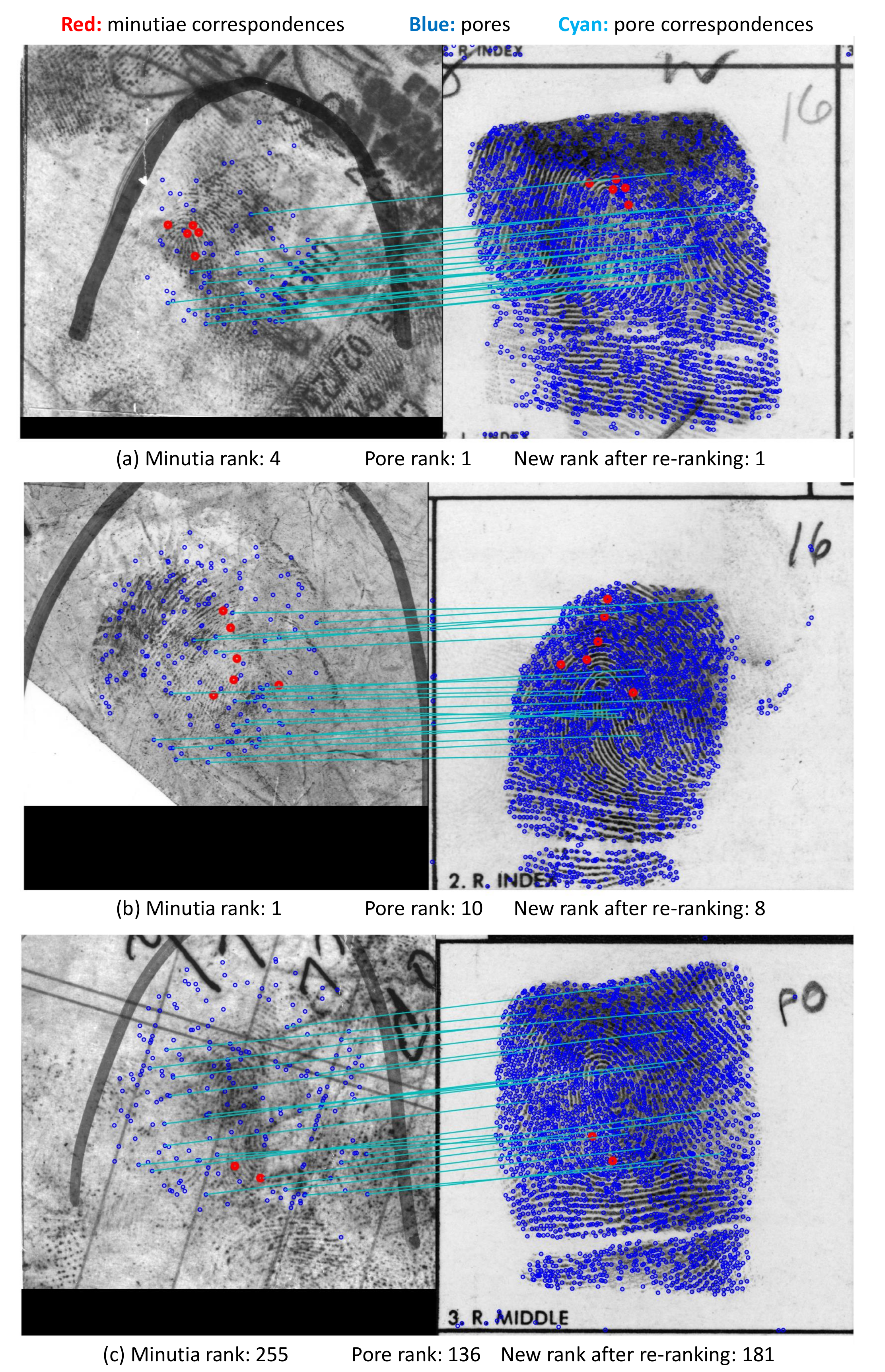}
\caption{Three situations, where the pore-based matcher: (a) helps minutiae matcher; (b) decreases performance of minutiae matcher; and (c) fails both before and after re-ranking.}
\label{fig:GoodMatch}
\end{figure}

Figure \ref{fig:GoodMatch} shows examples of correct and incorrect matches in NIST SD27, as well as situations where re-ranking does and does not improve the rankings for genuine mates.

\begin{figure*}[!tbp]
\centering
\includegraphics[width=\textwidth]{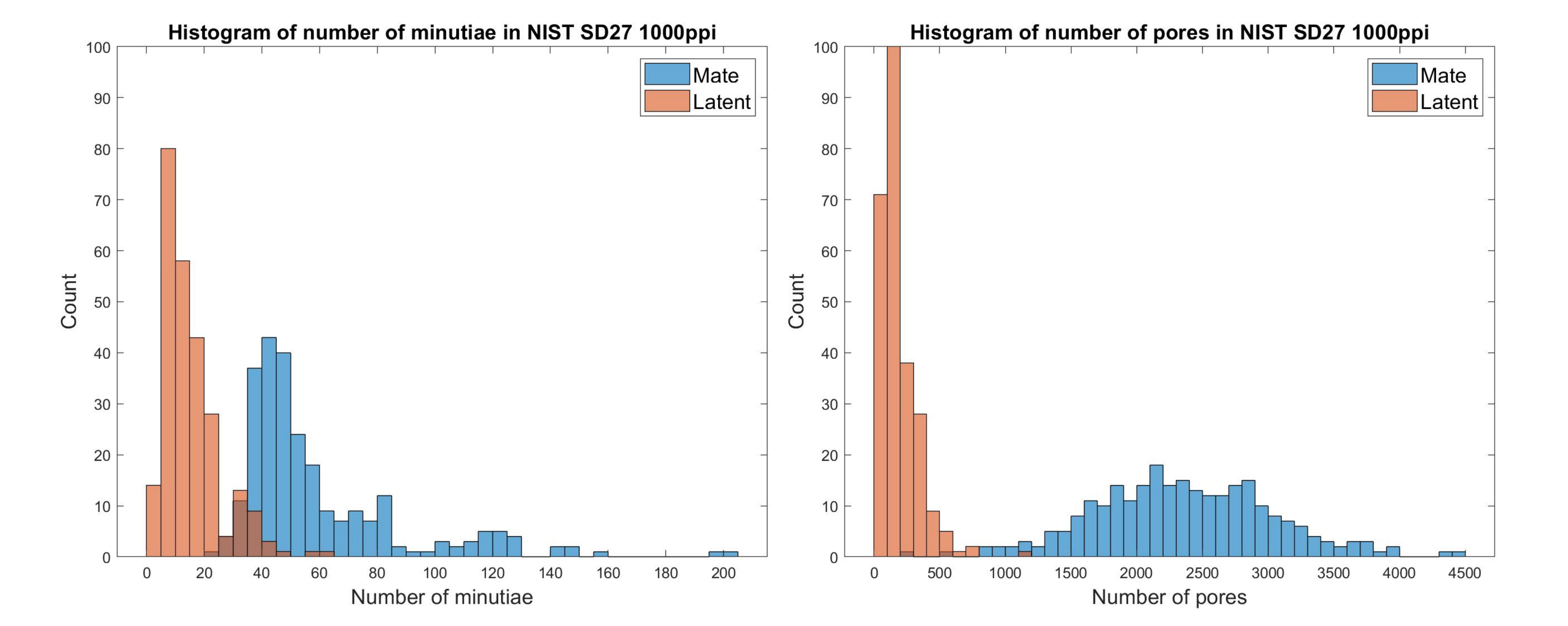}
\caption{Histograms of number minutiae and pores in both latent and rolled prints from NIST SD27 1,000 ppi dataset. The number of pores are significant larger than the number of minutiae.}
\label{fig:NumberMnt_Pores}
\end{figure*}

\begin{figure}[!tbp]
\centering
\includegraphics[width=\columnwidth]{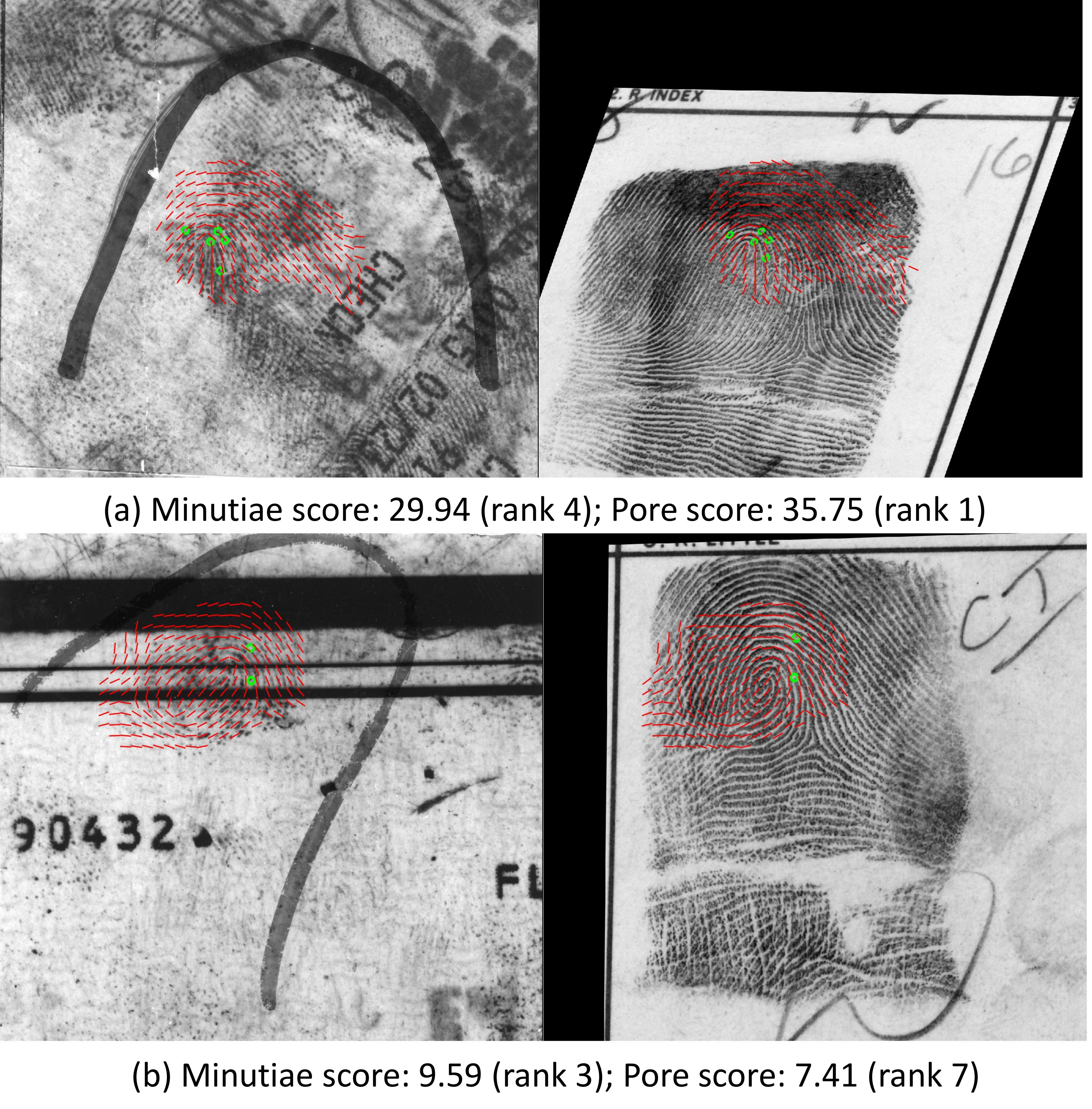}
\caption{Two examples of latent fingerprint distortion (left: latent image, right: its true mated print): (a) a case of minor distortion where the pore-based matcher helps improve the retrieval rank; and (b) large distortion that leads to incorrect ridge flow estimation and low pore and minutiae match scores. Latent image is overlaid on its true mate}
\label{fig:Overlay}
\end{figure}

We also evaluate our matcher on the NIST SD30 dataset. Table \ref{tab:NISTSD30} summarizes the results on the background size of $10,000$. Since this is a rolled print dataset, with high numbers of minutiae and low background noise, the minutiae-based matcher alone performs extremely well. As in Table \ref{tab:NISTSD30}, we clearly see that the pore based matcher helps improve the identification accuracy. 

\begin{table}[!tp]
	\centering
	\caption{Identification results on NIST SD30 dataset}    
	\label{tab:NISTSD30}
	\begin{small}
	\begin{adjustbox}{max width=\textwidth}
		\begin{tabular}{c|c|c|c}
			\hline
			Matcher&Rank-1&Rank-5&Rank-10\\
			\hline
			\hline
			{Minutiae only}	&$96.4\%$&$98.3\%$&$99.4\%$\\
			{Proposed approach}	&$98.1\%$&$99.1\%$&$100\%$\\
			\hline
		\end{tabular}
		\end{adjustbox}
	\end{small}
\end{table}

The current runtime of the pore-based matcher is, on average, 10 ms per match (matching one latent to one rolled print) on an Intel(R) Core(TM) i7-7700 CPU @3.6GHz, 32GB RAM desktop machine.

\subsection{Analysis of pore extraction and matching}
\label{sec:analysis}
In order to understand the performance and improve the recognition accuracy, we conducted ablation studies of our pore extraction and matching. Figure \ref{fig:NumberMnt_Pores} compares the number of minutiae and pores automatically extracted in both latent and mated prints in NIST SD27 1,000 ppi dataset. As confirmed by the figure, the partial nature of latent prints leads to a smaller number of minutiae and pores relative to rolled prints. Another observation is that the number of pores is significantly larger than number of minutiae ($5$ to $10$ times larger). While the number of pores in principle should help improve matching results by providing more reliable correspondences, they become very sensitive to noise if the pore distribution is too dense. This means we could find a match in a rolled print for nearly every pore in the latent print. Hence, the pore scores are not reliable.

Figure \ref{fig:Overlay} shows the overlay between latent and rolled prints with their corresponding rank. The greater the transformation  between the latent and its mated rolled print on the right, the more distortion it has. This distortion is one of the greatest challenges in the latent identification problem. This distortion is particularly difficult to handle in latent prints because there is insufficient surface area to capture reliable keypoints for rectification.

\subsection{Ablation study}
In this section, we further analyze the pore features on infant fingerprint besides adult fingerprints (NIST databases, PolyU-HRF, etc.), which have been intensively studied in literature. The infant fingerprint database \footnote{We contacted the authors of \cite{engelsma2019infant}} for our experiment contains $1,724$ images at $1,900$ ppi resolution from $194$ infants collected from $3$ sessions. For further details of the database, we refer readers to \cite{engelsma2019infant}.

\begin{figure}[!tbp]
\centering
\includegraphics[width=\columnwidth]{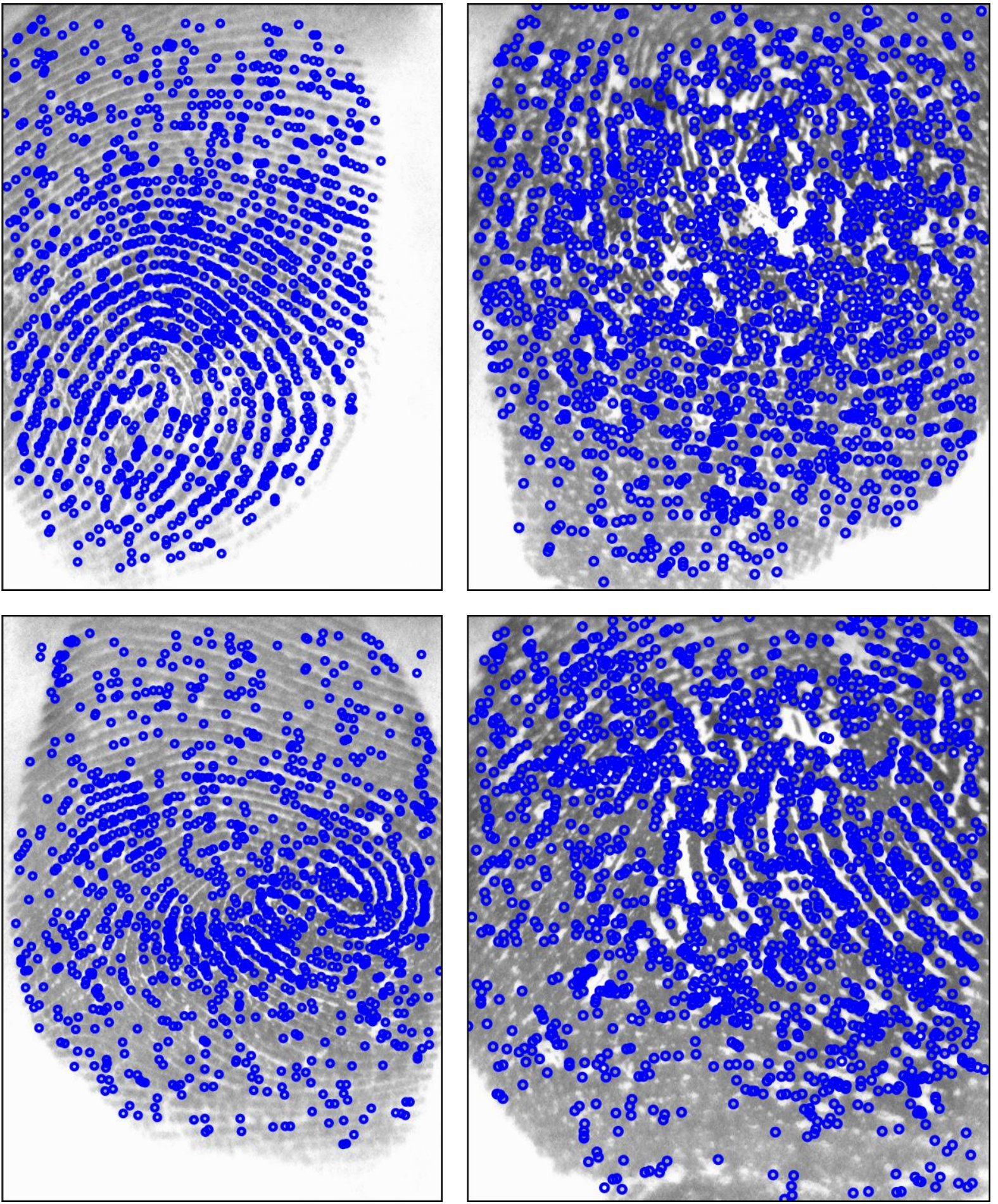}
\caption{Pore extraction on infant fingerprint images ($1,900$ ppi) with different qualities.}
\label{fig:InfantExtraction}
\end{figure}

Figure \ref{fig:InfantExtraction} shows the pore extraction on infant fingerprints. Although the pore size of infant prints are different from adult prints, the proposed pore extraction can also work well on those images. Note that we use the latent settings for both extraction and matching experiments for the infant database in this ablation study to demonstrate the portability of our approach. We follow the experimental setup of \cite{engelsma2019infant} in our verification experiment. Specifically, we used two state-of-the-art COTS matchers (COTS-A matcher is designed for latent fingerprint and COTS-B is utilized for plain (slap) prints), and the texture-based matcher from \cite{engelsma2019fingerprints}. The goal is to see whether using the scores of our pore matcher can help increase identification/verification accuracy. In our experiment, we fuse the scores of two thumbs of an infant and also scores of different impressions of the same thumb. Table \ref{tab:InfantMatching} shows the verification accuracy (TAR(\%)@0.1\% and 1.0\% FAR) of each matcher and the fusion of all matchers on $51$ 2-3 months old infants.

\begin{table*}[!tbp]
	\centering
	\caption{Verification accuracy (TAR(\%)@0.1\% and 1.0\% FAR) of different matchers (pore, COTS, and texture) on infant older than 2 months.}    
	\label{tab:InfantMatching}
	\begin{normalsize}
	\begin{adjustbox}{max width=\textwidth}
		\begin{tabular}{c|c|c|c|c|c|c|c}
			\toprule
			\multicolumn{2}{c|}{Fusion type}&Pores&COTS-A&COTS-B&Texture&COTS-A+B+Texture&Fusion all\\
			\hline
			\hline
			{All impressions}&{0.1\% FAR}	&$52.94\%$&$72.54\%$&$54.90\%$&$31.37\%$&\textbf{88.23\%}&$82.35\%$\\
			{}&{1.0\% FAR}	&$60.78\%$&$80.39\%$&$64.71\%$&$52.94\%$&\textbf{94.11\%}&\textbf{94.11\%}\\
			\bottomrule
		\end{tabular}
		\end{adjustbox}
		\end{normalsize}
\end{table*}

\begin{figure}[!tbp]
\centering
\includegraphics[width=\columnwidth]{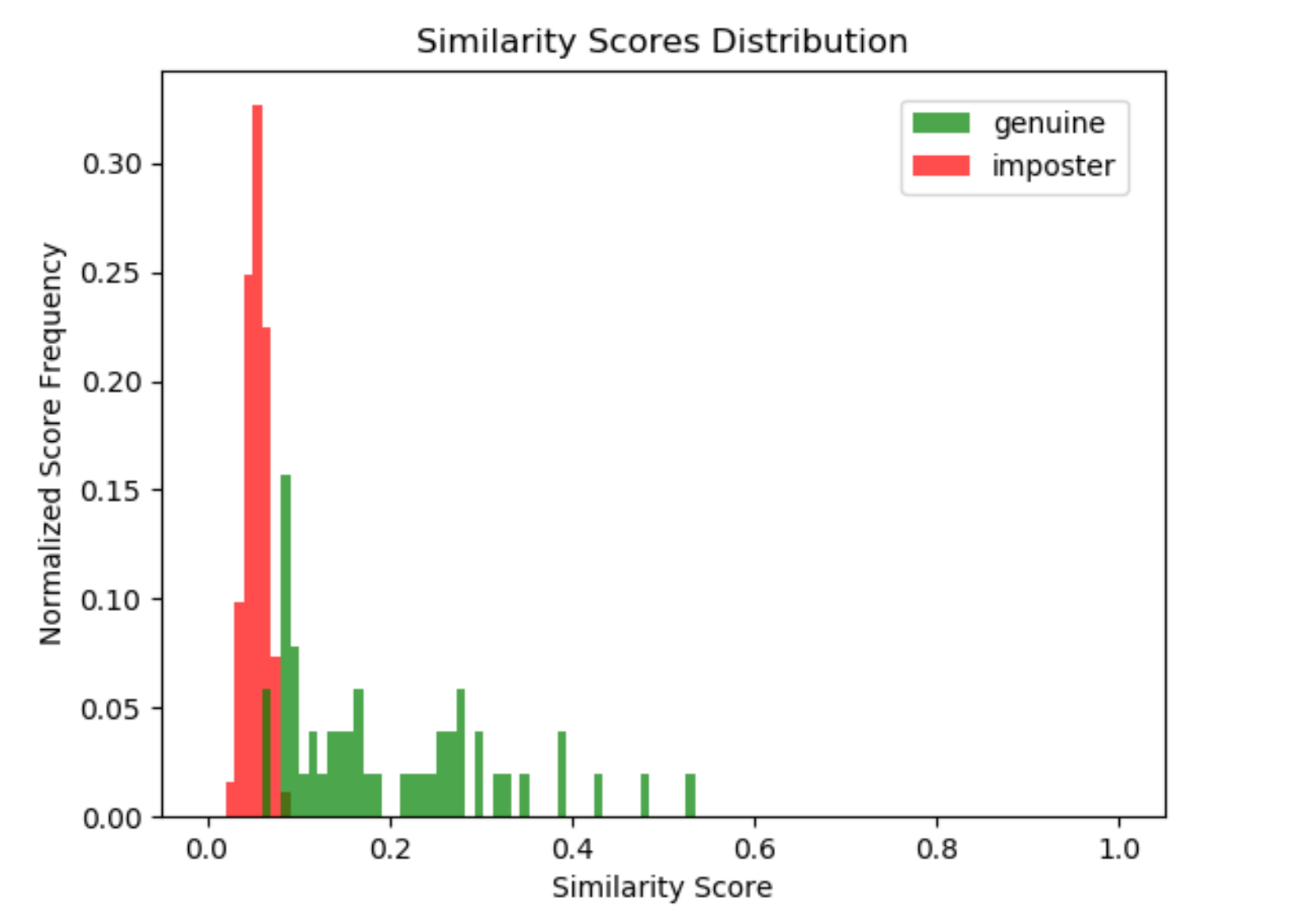}
\caption{Distribution of the genuine and imposter scores of infant fingerprints by fusing all impressions. Simple score fusion of different matchers (pores, COTS, and texture) can separate those distributions.}
\label{fig:InfantHistogram}
\end{figure}

As from the table, although the individual pore scores do not have the high TAR values compared to the two COTS matchers, which utilize minutiae information, fusing all scores together helps boost the performance. Figure \ref{fig:InfantHistogram} shows the genuine and imposter score histogram of infant fingerprints. Even though the distribution of imposter scores has a long tail, score level fusion between matchers helps separate those genuine and imposter scores.

\section{Conclusion}
In this paper, we have presented an end-to-end pore extraction and matching framework for arbitrary fingerprint types and qualities. To demonstrate the effectiveness of our proposed approach, we evaluated our approach in terms of qualitative, quantitative, and recognition accuracy. Experimental results and analysis reveal that including pore based matcher leads to only a marginal improvement in retrieval results compared to minutiae only based matcher. This can be explained in terms of large distortion in latent prints where pores are not clearly visible. However, pores do improve in some latents where the number of minutiae is neither too large nor too small. So, this is some evidence of the complementary characteristics of pores and minutiae, but generalization of these cases is difficult. The pore matcher could be further improved in the following ways: (i) improve fingerprint normalization steps to compensate for fingerprint distortion; and (ii) incorporate additional features such as ridge flow for the re-ranking step. Currently, the only 1,000 ppi latent database is the NIST SD27 with 255 image pairs. Lack of additional high resolution latent datasets (with associated mates and a large background) is a major stumbling block.


%

%


%

\ifCLASSOPTIONcaptionsoff
  \newpage
\fi



%

%
%

\bibliographystyle{ieee}
\bibliography{Journal_ref}

%

%
%
%




\end{document}